\documentclass{IOS-Book-Article}

\usepackage{mathptmx}
\usepackage[table]{xcolor}
\usepackage{hyperref}
\usepackage{graphicx}
\usepackage{easyReview}

\def\hb{\hbox to 10.7 cm{}}

\begin{document}

\pagestyle{headings}
\def\thepage{}

\begin{frontmatter}              

\title{Computer-Assisted Creation of Boolean Search Rules for Text Classification in the Legal Domain}

\markboth{}{December 2019\hb}

\author[A]{\fnms{Hannes} \snm{Westermann},%
\thanks{Corresponding Author: Hannes Westermann, E-mail: hannes.westermann@umontreal.ca}},
\author[B]{\fnms{Jarom\'{i}r} \snm{\v{S}avelka}},
\author[C]{\fnms{Vern R.} \snm{Walker}},
\author[B]{\fnms{Kevin D.} \snm{Ashley}} and
\author[A]{\fnms{Karim} \snm{Benyekhlef}}

\runningauthor{H. Westermann et al.}
\address[A]{Cyberjustice Laboratory, Facult\'e de droit, Universit\'e de Montréal}
\address[B]{ISP, School of Computing and Information, University of Pittsburgh}
\address[C]{LLT Lab, Maurice A. Deane School of Law, Hofstra University}

\begin{abstract}
In this paper, we present a method of building strong, explainable classifiers in the form of Boolean search rules. We developed an interactive environment called CASE (Computer Assisted Semantic Exploration) which exploits word co-occurrence to guide human annotators in selection of relevant search terms. The system seamlessly facilitates iterative evaluation and improvement of the classification rules. The process enables the human annotators to leverage the benefits of statistical information while incorporating their expert intuition into the creation of such rules. We evaluate classifiers created with our CASE system on 4 datasets, and compare the results to machine learning methods, including SKOPE rules, Random forest, Support Vector Machine, and fastText classifiers. The results drive the discussion on trade-offs between superior compactness, simplicity, and intuitiveness of the Boolean search rules versus the better performance of state-of-the-art machine learning models for text classification. 
\end{abstract}

\begin{keyword}
Artificial Intelligence \& Law\sep Text Classification\sep Semantic exploration \sep 
Boolean search \sep Natural language processing \sep Explainable artificial intelligence
\end{keyword}
\end{frontmatter}
\markboth{December 2019\hb}{December 2019\hb}

\section{Introduction}

Reading, interpreting, and understanding legal texts is one of the most important skills of legal professionals. Lawyers, judges, students, and researchers alike spend a lot of time and effort learning, honing, and using the skill in reading statutory law, a legal case, a contract, or a journal article, while interpreting the document and applying the knowledge to solve a new problem. Such analysis is done on several levels---sometimes, an individual sentence carries the much needed information, while other times the reader has to study whole sections or the entire document to understand the important point. Further, the reader may have to understand different features, such as the facts of a legal case or the relevance of a sentence.

The ability to categorize the texts or their pieces into certain types (e.g., court reasoning, legal rule, facts) is an integral part of the analysis. It is therefore no coincidence that text classification is one of the big focus areas in the field of artificial intelligence and law (AI \& Law). Automating the classification tasks has often become feasible due to advances in machine learning (ML) and natural language processing (NLP) methods. Typically, the research is conducted by  manually labelling hundreds or thousands of documents. Once the annotation is completed, the researchers use the data to build ML models that are able to learn patterns in the annotated data and apply these to classify  new unseen texts.

An automated approach has a host of advantages when compared to the tedious work of classifying the whole corpus without the help of a computer. However, there are some drawbacks as well. Firstly, it still takes a lot of effort to manually label the subset of documents required for  training  an ML classifier. Secondly, it can be difficult to explain the decisions of sophisticated models. This may sometimes lead to skepticism as to the suitability of such models to be used in practice. Possible over-fitting is yet another risk which needs to be taken into account. Sometimes the models work well on the annotated data, but fail to generalize to unseen documents. 

In this paper, we present a method addressing these issues. We built a tool that allows annotators to create Boolean rules in a computer-assisted fashion. These rules could potentially be used for classification in domains with little available data, by incorporating human intuition into the process. Further, the rules created are more explainable than most machine learning models, while still performing reasonably well. 

\section{Prior work}

According to Antonie and Zaiane \cite{antonie:2002} ``a good text classifier [\ldots] efficiently categorizes large sets of text documents in a reasonable time frame and with an acceptable accuracy, and [\ldots] provides classification rules that are human readable for possible fine-tuning.'' One approach to text classification is to let a human expert define a set of logical rules based on his domain-specific knowledge of how to classify documents under a given set of categories \cite{korde:2012}. Generating rules based on human expertise is time-consuming, expensive, and sometimes not feasible. However, the great advantage of such rules is that they often provide intuitive and meaningful explanation (justification) of the resulting classification.

Alternatively, one can apply various methods for inducting text classification rules automatically including such methods as decision trees or associative rule mining \cite{korde:2012}. The latter employs an iterative search of a database to discover the most frequent sets of k items (k-itemsets) that are associated with the documents sharing a particular classification; a logical rule based on a k-itemset should support the classification with a confidence above a certain threshold. The potentially very large number of rules are then pruned using various techniques \cite{antonie:2002}. A disadvantage of such automatically learned rules is that they may not correspond to expert intuitions about texts in the domain.

Various hybrids of manual and automated methods are possible. For example, Yao, et al. \cite{yao:2019} evaluated a medical clinical text classification method that employed rules to identify trigger phrases such as disease names and alternatives. They used the trigger phrases to predict classes that had very few examples. For the remaining classes they trained a knowledge-guided convolutional neural network (CNN) with word embeddings and medical feature embeddings.

In Walker et al. \cite{walker:2019}, the authors investigated the task of automatically classifying, within adjudicatory decisions in the United States, those sentences that state whether the conditions of applicable legal rules have been satisfied or not (``Finding Sentences"), by analyzing a small sample of classified sentences (N = 530) to manually develop rule-based scripts, using semantic attribution theory. The methodology and results suggested that some access-to-justice use cases can be adequately addressed at much lower cost than previously believed. Our work extends that effort by developing a platform for efficiently improving the classification rules in the iterative fashion.

\section{Boolean Search Rules}

We propose and evaluate a novel hybrid combination of manual and automated construction of text classification rules. Our CASE system helps annotators select relevant terms, create Boolean text classification rules, and evaluate and improve them in an iterative manner. Depending on the use case, the resulting rules may prove very useful---especially where explanatory power and compactness are important.

``Boolean search rules'' are an appealing method for classifying documents because such rules are familiar to anyone who works with legal information retrieval systems. They make it possible to search for single words (such as ``veteran"), which would return all cases containing the word. Further, it is possible to logically combine several rules, using the OR, AND, and NOT operators. OR returns texts with either of the two words while AND requires both of them to be present. NOT excludes texts containing a particular word. In our case, we are using the FTS5 search engine integrated into the SQLite Database \cite{SQLite} to process our queries. This allows us to build complex queries, combining different logical operators, that are executed very rapidly.

\subsection{Existing methodologies to create Boolean rules}
There have been previous attempts of using Boolean search rules in AI \& Law. However, without the methods presented in this paper the process can be long and laborious. A recent attempt at creating such search rules was made by Walker et al. \cite{walker:2019}. The researchers tested whether distinctive phrasing in legal decisions enables the development of automatic classifiers on the basis of a small sample of labeled decisions, with adequate results for some important use cases. Certain words, such as ``finds", were found to closely correspond to a sentence having the rhetorical role of a finding of fact. Two such rules were tested, leading to an F1 score of 0.512 in identifying such sentences.
Testing new hypotheses, observing the results, and comparing the results of new classification rules against the old ones, was a time-consuming and laborious process. In this paper, we introduce a tool that makes such a process  more efficient.

\section{Methodology}

In this paper, we test the hypothesis that Boolean search rules created by humans with the assistance of a computerized tool can prove useful in building text classifiers in the legal domain. To test the hypothesis we created such rules on four datasets of case texts, and compared the results to those obtained by using ML methods. The process is described in this section.

\subsection{Datasets}
\label{sec:datasets}
We selected four existing datasets created within the AI \& Law community to evaluate our methodology. These are presented below.

\subsubsection{Veterans Claims Dataset (sentence roles)}
Walker et al. \cite{walker:2019} analyzed 50 fact-finding decisions issued by the U.S. Board of Veterans' Appeals (``BVA") from 2013 through 2017, all arbitrarily selected cases dealing with claims by veterans for service-related post-traumatic stress disorder (PTSD). For each of the 50 BVA decisions in the PTSD dataset, the researchers extracted all sentences addressing the factual issues related to the claim for PTSD, or for a closely-related psychiatric disorder. These were tagged with the rhetorical roles \cite{walker:2017} the sentences play in the decision. We conducted our experiments on this set of sentences.

\subsubsection{Court Decisions Segmentation Dataset (functional parts)}
\v{S}avelka and Ashley \cite{savelka:2018} examined the possibility of automatically segmenting court opinions into high-level functional parts (i.e., Introduction (I), Background (B), Analysis (A), Footnotes (F)) and issue specific parts (i.e., Conclusions(C)). They assembled 316 court decisions from Court Listener and Google Scholar, 143 in the area of cyber crime and 173 involving trade secrets. These were annotated, after which Conditional Random Fields (CRF) models were trained to recognize the boundaries between the sections. We used the cases in the area of cyber crime for our tests. It should be noted that we do not attempt to detect the boundaries, but instead try to classify the annotated text sections.

\subsubsection{The Trade Secrets Factors Dataset (factor prediction)}
Falakmasir and Ashley \cite{falakmasir:2017} assembled a corpus of 172 trade secret misappropriation cases employed in the HYPO, CATO, SMILE+IBP and VJAP programs.  Legal experts
had labeled the cases by the applicable factors, stereotypical patterns of fact that strengthen or weaken a claim. There are 26 trade secret misappropriation factors. For our experiments, we used the existence of security measures in a case (Factor 6), to deal with a binary classification task.

\subsubsection{The Statutory Interpretation Dataset (interpretative value of sentences}
\v{S}avelka et al. \cite{savelka:improving} studied methods for retrieving useful sentences from court opinions that elaborate on the meaning of a vague statutory term. To support their experiments they queried the database of sentences from case law that mentioned three terms from different provisions of the U.S. Code. They manually classified the sentences in terms of four categories with respect to their usefulness for the interpretation of the corresponding statutory term. Here we work with the sentences mentioning `common business purpose' (149 high value, 88 certain value, 369 potential value, 274 no value). In \cite{savelka:improving} the goal was to rank the sentences with respect to their usefulness; here, we classify them into the four value categories.

\subsection{Dataset Split}
The four datasets are described in section \ref{sec:datasets}. For our experiments we split each dataset into three parts:  training (20\%), validation (10\%), and testing (70\%). The unusual split (small training set) was used to evaluate the performance of the search queries in situations where very little data is available. This is often the case in the problems of interest in the field of AI \& Law. Using the identical dataset splits we created classifiers with the CASE tool and ML methods, as described below.

\subsection{CASE - Computer Assisted Semantic Exploration}

\begin{figure}
  \includegraphics[ trim={0 0 0 0},width=\linewidth]{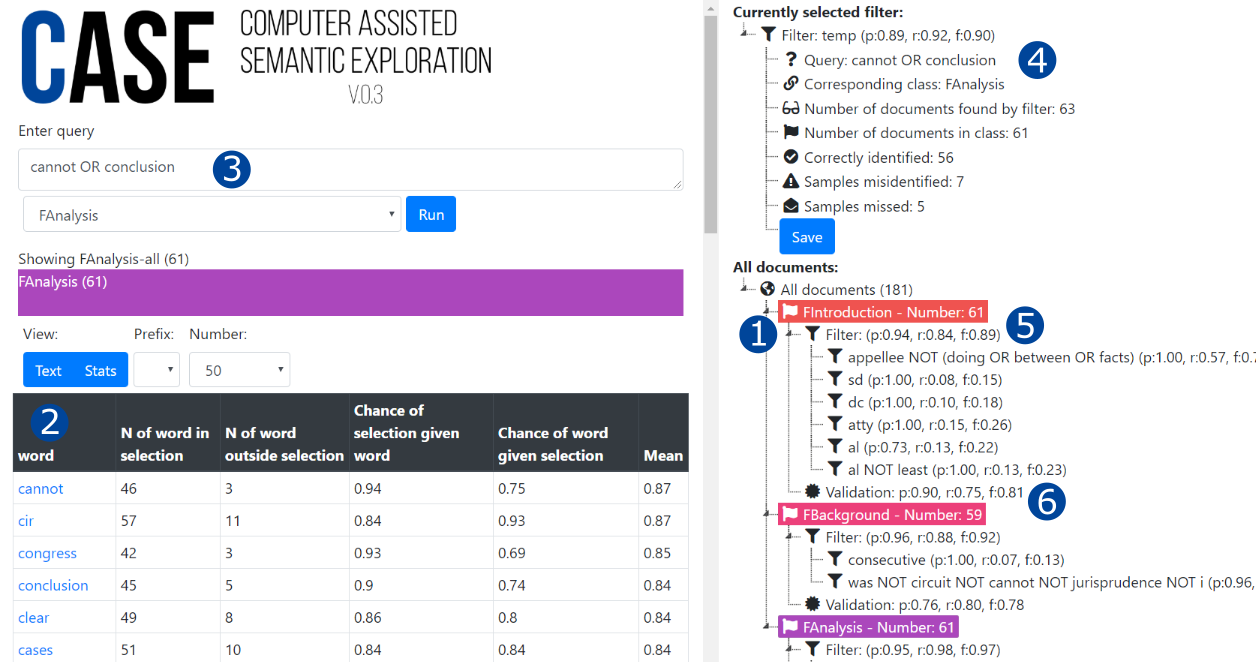}
  \caption{The Computer Assisted Semantic Exploration (CASE) interface.}
  \label{fig:case}
\end{figure}

We developed a tool for Computer Assisted Semantic Exploration (CASE). CASE facilitates seamless creation of Boolean classification rules. Figure \ref{fig:case} shows a screenshot of the interface. The tool supports users in interactively creating Boolean rules using several statistical methods. At (1), CASE displays the possible classes for annotation. By clicking on a class, the user selects texts in that specific class, and is then shown information about the word distribution inside that selection under (2). This list is sortable, and shows several headers, containing metrics useful for the selection of significant words.

Once a user has found a word that is a strong indicator of a specific class, he can create a query in (3), using logical operators such as AND, OR, and NOT. For example, a query to identify the class Analysis could be ``cannot OR conclusion.'' The query can then be run, and is immediately evaluated, with the results being presented in (4). Here, the user also has the possibility of selecting documents that are misidentified, for example, in order to exclude certain words. The user can thus work on creating queries able to identify classes with high precision and recall in an iterative fashion. 

Once the user is content with a query, he can save it and create additional queries. Ideally, in conjunction, the queries will identify documents with high precision and recall. The filters are constantly evaluated against the training data (5), and validation data (6) to prevent over-fitting.

The queries used for the current paper were created by the co-authors of this paper. Using the statistics provided by CASE as well as our previous intuitions about the datasets, we used the tool to add and modify rules until it was difficult to introduce new rules without lowering the validation score. CASE was very helpful in identifying words significant for a class and using these to create the rules.

\subsection{Machine Learning}


We trained four different types of ML models as benchmarks. We used SKOPE-rules \cite{skope} to simulate the situation where the model is forced to construct similar Boolean rules as a human using CASE. The difference is that the rules are learned automatically. We trained random forest classifier, support vector machine (SVM), and fastText \cite{joulin:bag} models on more sophisticated features. Their predictions are used to investigate how much performance one has to sacrifice in order to benefit from the explanatory power of CASE (computer assisted) and SKOPE-rules (computer generated). We have used the same training sets as those in the CASE experiments to train the models. The validation sets were used to optimize the models' hyperparamters. The same test sets were used for the evaluation.

SKOPE-rules is a Python ML module the aim of which is learning logical, interpretable rules. \cite{skope} A decision rule is a logical expression of the form ``IF conditions THEN response.'' The problem of generating such rules has been widely considered in ML, see e.g., RuleFit \cite{friedman:predictive}, Slipper \cite{cohen:simple}, LRI \cite{weiss:lightweight}, MLRules \cite{dembczynski:maximum}. SKOPE-Rules extracts rules from an ensemble of trees. A weighted combination of these rules is then built by solving an L1-regularized optimization problem over the weights as described in \cite{friedman:gradient}. To force the model to construct the rules that are comparable to those created using CASE, we have used unigram, bigram, and trigram word occurrences as features. The classification model is then a set of rules (possibly overlapping, i.e., OR), where each rule is a conjunction (i.e., AND) of matching or filtering (NOT) on words and phrases. For each data set we have trained a number of binary models, one for each class. 

A random forest is an ensemble classifier that fits a number of decision trees on sub-samples of the data set. It uses averaging to improve the predictive accuracy and control over-fitting. As an implementation of random forest we used the scikit-learn's Random Forest Classifier module \cite{rfcm}. As features we use TF-IDF weights of (1-4)-grams of lowercase tokens with their POS tags. 

An SVM classifier constructs a hyper-plane in a high dimensional space, which is used to separate the classes from each other. As an implementation of SVM we used the scikit-learn's Support Vector Classification module \cite{svc}. We used the same features as with the random forest models to train a number of binary classifiers. 

FastText is a linear classifier that uses ngram features that are embedded and averaged to form the hidden variable. We worked with the Python wrapper \cite{fasttext:pwrap} for the original library released by Facebook \cite{fasttext:lib}. As for other classifiers we trained a number of binary classification models using grid search to optimize hyperparameters.

\subsubsection{Evaluation}
The evaluation of all the models is performed on the test sets (70\% of the respective datasets). Note that all the methods were trained on the identical training sets and fine-tuned on the identical validation sets. The performance is measured in terms of precision (P), recall (R), and F$_1$-measure (F$_1$). All the classifiers are evaluated in the one-vs-rest settings where each label within each of the four datasets has its own classifier. We measure aggregate results as well. ``Overall" averages the scores for the different classes over the total number of classes. ``Overall-w" uses a weighted average, where each class is given a weight according to how often it appears in the test dataset. For each dataset, the highest overall F$_1$-score is written in bold.

\section{Results}
\label{sec:results}
The results are presented in Table \ref{tab:scores}. In general, the rules created by CASE   performed similarly to the computer generated SKOPE rules. However, they seem to have a slightly higher precision, with a lower recall. This can have utility for certain use cases. As expected the more complex RF, SVM, and fastText models perform better than the human generated rules. We discuss the trade-offs between explainability and performance below.

\begin{table}[]
    \centering
    \setlength{\tabcolsep}{5pt}
    \def\arraystretch{1.1}
    \begin{tabular}{l|rrr|rrr|rrr|rrr|rrr}
                   & \multicolumn{3}{c|}{CASE} & \multicolumn{3}{c|}{SKOPE} & \multicolumn{3}{c|}{RF} & \multicolumn{3}{c|}{SVM} & \multicolumn{3}{c}{Fasttext}\\
                   & P   & R   &F$_1$& P   & R   &F$_1$& P   & R   &F$_1$& P   & R   &F$_1$& P   & R   &F$_1$ \\
        \multicolumn{16}{c}{\cellcolor{black!8}VetClaims}     \\
        -sentence  & .84 & .25 & .38 & .80 & .33 & .47 & .90 & .61 & .72 & .99 & .44 & .61 & .87 & .61 & .72  \\
        -finding   & .71 & .38 & .50 & .63 & .40 & .49 & .77 & .26 & .39 & .83 & .57 & .67 & .68 & .59 & .63  \\
        -evidence  & .82 & .74 & .78 & .71 & .81 & .76 & .88 & .88 & .88 & .90 & .92 & .91 & .87 & .92 & .90  \\
        -rule      & .71 & .48 & .57 & .60 & .65 & .63 & .95 & .55 & .70 & .90 & .78 & .83 & .87 & .79 & .82  \\
        -citation  & .96 & .99 & .97 & .86 & .86 & .86 & .99 & .96 & .97 & .98 & .98 & .98 & .99 & .97 & .98  \\
        -reasoning & .62 & .14 & .22 & .50 & .23 & .31 & .65 & .06 & .12 & .75 & .27 & .39 & .43 & .39 & .41  \\
        -overall   & .78 & .50 & .57 & .68 & .55 & .59 & .86 & .55 & .63 & .89 & .66 & .73 & .79 & .71 & \textbf{.74}  \\
        -overall-w & .80 & .61 & .67 & .74 & .71 & .71 & .88 & .68 & .73 & .96 & .83 & \textbf{.87} & .83 & .80 & .81  \\
        \multicolumn{16}{c}{\cellcolor{black!8}Section segmentation}     \\
        -intro     & .90 & .75 & .81 & .83 & 1.0 & .91 & 1.0 & .98 & .99 & 1.0 & .99 & .99 & .98 & .95 & .97  \\
        -backg.    & .76 & .80 & .78 & .62 & .96 & .75 & .97 & .83 & .90 & .99 & .85 & .91 & .96 & .85 & .90  \\
        -analysis  & .87 & .83 & .85 & .88 & .88 & .88 & .98 & .88 & .93 & .93 & .97 & .95 & .90 & .98 & .94  \\
        -overall   & .84 & .79 & .81 & .78 & .95 & .85 & .98 & .90 & .94 & .97 & .94 & \textbf{.95} & .95 & .93 & .94  \\
        -overall-w & .84 & .79 & .82 & .78 & .95 & .85 & .98 & .89 & .94 & .97 & .94 & \textbf{.95} & .95 & .93 & .94  \\
        \multicolumn{16}{c}{\cellcolor{black!8}Trade secrets}        \\
        -security  & .65 & .61 & .63 & .53 & .97 & \textbf{.69} & .59 & .69 & .64 & .50 & 1.0 & .67 & .57 & .49 & .53  \\
        \multicolumn{16}{c}{\cellcolor{black!8}Statutory interpretation}    \\
        -high      & .72 & .45 & .55 & .66 & .39 & .49 & .91 & .10 & .17 & .96 & .22 & .36 & .61 & .46 & .52  \\
        -certain   & .18 & .18 & .18 & .26 & .23 & .24 & .67 & .13 & .22 & .60 & .10 & .17 & .40 & .13 & .20  \\
        -potential & .69 & .36 & .47 & .49 & .98 & .65 & .69 & .54 & .60 & .71 & .64 & .67 & .63 & .68 & .65  \\
        -no        & .89 & .65 & .75 & .74 & .71 & .73 & .90 & .77 & .83 & .90 & .78 & .83 & .92 & .79 & .85  \\
        -overall   & .62 & .41 & .49 & .54 & .58 & .53 & .79 & .39 & .45 & .79 & .44 & .51 & .64 & .52 & \textbf{.55}  \\
        -overall-w & .70 & .44 & .54 & .57 & .72 & .61 & .79 & .50 & .56 & .80 & .56 & .62 & .69 & .62 & \textbf{.64}  \\
    \end{tabular}
    \caption{P, R and $F_1$ for the different classifiers applied on datasets described in Section \ref{sec:datasets}}.
    \label{tab:scores}
\end{table}

\textbf{Veteran Claims Dataset.}
Compared to the work in \cite{walker:2019}, the CASE tool gave us significant flexibility and speed improvements in creating and optimizing the Boolean search rules. For ``finding sentences,'' for example, we confirmed the usefulness of the ``finds" and ``preponderance" search terms \cite{walker:2019}, while adding others such as ``elements" and ``warranted".

\textbf{Court Decisions Segmentation Dataset.}
The CASE rules achieved higher precision, but lower recall than the SKOPE rules. The created Boolean search rules are quite simple. For identifying the analysis section, for example, the following query was quite successful: ``cannot OR apparent OR definition OR prohibition.'' 

\textbf{Trade Secrets Factors Dataset.}
This dataset was the most difficult to deal with. There were few cases, and they were long and complex. For training, only 33 cases were available.  However, the rules achieved the highest precision among the classifiers. We relied heavily on human intuition, such as the term ``non-disclosure'' implying the existence of security measures. Building the rules also helped us identify an error in the annotation of a case.

\textbf{Statutory Interpretation Dataset.}
This dataset was also very hard to deal with, due to its being unbalanced and the fact that the value of a sentence for statutory interpretation is hard to link to individual terms. Again, we can see the pattern of the CASE rules having higher precision than SKOPE rules, but lower recall.

\begin{table}[]
    \footnotesize
    \renewcommand{\arraystretch}{1}
    \centering
    \begin{tabular}{p{3.6cm}p{3.6cm}p{3.6cm}}
        CASE (f1: .85)  & SKOPE-rules (f1: .88)                  & Random Forest (f1: .93) \\
        (boolean rules) & (boolean rules)                        & (important features) \\
        \hline
        cannot OR        & (cir AND NOT headquarters in           & is not, that, be, cir, cases, can, \\
        apparent OR        & AND is) OR                               & is, provides, held, it, statute, in, \\
        prohibit OR      & (2d AND NOT february 17                & intended, record, see also, there, \\
        definition      & AND is not) OR                           & 9th cir, subsection, 7th cir, of  \\
                        & (NOT appeal AND NOT cir                & such, have, may be, congress,  \\
                        & AND it is) OR                             & when, issue, 3d at, evidence  \\
                        & (NOT cir AND is not AND                & that, if, thus the, as, where, \\
                        & NOT of 18)                             & is to, here the, definition of \\
    \end{tabular}
    \caption{Comparison between created rules for identifying the analysis section in the segmentation dataset.}
    \label{tab:comparison}
\end{table}

\subsection{Explainability of Rules or Features}

Table \ref{tab:comparison} shows a comparison of rules created using the different systems for classification of the ``analysis part'' of the court decisions segmentation dataset. For the CASE and SKOPE rules, a document triggering any of the listed queries will result in the document being labeled as ``analysis.'' For the random forest algorithm, we present the most important features, as selected by the algorithm. Overall, the CASE rules are much less complex, while still showing performance that is not much inferior. Further, the CASE rules seem to contain more legally relevant terms, such as ``prohibit'' and ``definition.'' These properties make the rules easier to explain.

\section{Discussion}

We have shown that Boolean search rules can be created efficiently with a system such as CASE. In most areas, the performance was weaker than their ML counterparts. However, the CASE rules have advantages that might make their use desirable in some use cases. In this section, we discuss some of the advantages and disadvantages of using such Boolean search rules for classification in legal domains.

\subsection{Advantages of using Boolean rules}
One advantage of using Boolean rules, developed with the assistance of the CASE platform, is that those rules can \textit{incorporate human intuitions}. Thus, the user can rapidly formulate and evaluate hypotheses of which terms might prove useful in search rules, assisted by the statistical measures provided by CASE. In doing so,  users are able to select words that they know have legal significance in relation to the classifier.  

In incorporating this intuition, the user has significantly more \textit{control} over the created model than with ML systems. With ML it is difficult to direct the training of the system, beyond feature selection and hyperparameter optimization. In CASE, on the other hand, the human is always in control of the system. Evaluation occurs continuously, and the human has complete control over how the model develops and how new search terms affect the precision and recall of the search rules. This allows  users to fit the rules more exactly to their requirements and use case. Further, the creation process allows the annotator to develop an intuition for the particularities of the dataset in an exploratory fashion. In the trade secrets dataset, the system  helped us to discover an error in classification, showcasing this advantage.

The incorporation of human intuition, together with the level of control a human is given over the creation of the search rules, can potentially allow the user to create search rules that are less prone to overfit and therefore \textit{generalize} better. The user can choose to use only  phrases that are independent of the specific context of the dataset, thereby creating rules that generalize to other datasets. Since the users  decide whether to use  a term, even very small datasets 
could support the creation of the rules with high precision.

A big issue in the practical use of ML in the legal field is the difficulty of \textit{explaining} the created models. This might cause legal professionals not to trust the algorithms. Using Boolean search rules might alleviate this issue. Firstly, the human who makes the decisions in creating models, is fully aware of why a particular word was chosen and used in a certain way. Further, the structure of the created rules, using AND, NOT, and OR, should be easier to grasp than complex ML models. They can thus offer a basis for better explaining why a particular document was chosen, and why not. As can be seen in Table \ref{tab:comparison}, the CASE rules are both simpler and more legally relevant than the ML models.

\subsection{Limitations}
As can be seen from the results presented in Section \ref{sec:results}, ML models often performed better than the human-created rules. This is an interesting result in itself, as it shows the power of well-optimized ML methods even on small datasets. If performance is the most important metric, using ML methods could thus often be preferable. We discuss methods to combine advantages from ML and CASE below (Section \ref{sec:future}).

\section{Future work}
\label{sec:future}
This paper is an initial step in exploring the use of computer-assisted creation of 
Boolean search rules for text classification. There are many avenues for further research. One is to  expand the CASE system. For example, the system could include n-grams beyond simple words. Restructuring  the classifiers as multi-label classifiers, and running the best classifiers first, would  improve performance. The system should also be expanded to work better with long documents, such as the trade secrets cases. Another avenue is combining the CASE platform with ML methods in a hybrid approach to harness the advantages of both. For example, CASE could be used to preselect documents from a massive corpus, after which a ML algorithm could be trained on only those documents. Another approach would be to run a ML model on annotated data, and use CASE subsequently to analyze the results and exclude false positives.

\section{Conclusions}
In this paper we have proposed and evaluated CASE, a novel approach for computer-assisted text classification using Boolean matching rules. We have shown that in a number of use cases the rules perform surprisingly well using little annotated data while offering superior explanatory power when compared to ML methods.

\end{document}